\documentclass[conference]{IEEEtran}
\newcommand{\CLASSINPUTtoptextmargin}{1in}
\newcommand{\CLASSINPUTbottomtextmargin}{1in} 
\IEEEoverridecommandlockouts
\usepackage{cite}
\usepackage{amsmath,amssymb,amsfonts}
\usepackage{algorithm,algorithmic}
\usepackage{algorithmic}
\usepackage{graphicx}
\usepackage{textcomp}
\usepackage{xcolor}
\usepackage{lineno}
\usepackage{mathtools}
\usepackage{csquotes}

\def\BibTeX{{\rm B\kern-.05em{\sc i\kern-.025em b}\kern-.08em
    T\kern-.1667em\lower.7ex\hbox{E}\kern-.125emX}}

\makeatletter
\newcommand*\titleheader[1]{\gdef\@titleheader{#1}}
\AtBeginDocument{%
	\let\st@red@title\@title
	\def\@title{%
		\bgroup\normalfont\large\centering\@titleheader\par\egroup
		\vskip1.5em\st@red@title}
}
\makeatother

\begin{document}
\title{
An Explainable Artificial Intelligence Framework for Quality-Aware IoE Service Delivery
\thanks{Accepted article by IEEE International Conference on Communications (ICC 2022), ©2022 IEEE.}
\thanks{*Dr. CS Hong is the corresponding author.}
\thanks{©2022 IEEE. Personal use of this material is permitted. Permission from
	IEEE must be obtained for all other uses, in any current or future media,
	including reprinting/republishing this material for advertising or promotional
	purposes, creating new collective works, for resale or redistribution to servers
	or lists, or reuse of any copyrighted component of this work in other works.}
}
\author{\IEEEauthorblockN{Md. Shirajum Munir, Seong-Bae Park, and Choong Seon Hong*}
	\IEEEauthorblockA{\textit{Department of Computer Science and Engineering, Kyung Hee University, } \\
		\textit{Yongin-si 17104, Republic of Korea}\\
		E-mail: munir@khu.ac.kr, sbpark71@khu.ac.kr, cshong@khu.ac.kr}
}

\maketitle

\begin{abstract}
One of the core envisions of the sixth-generation (6G) wireless networks is to accumulate artificial intelligence (AI) for autonomous controlling of the Internet of Everything (IoE). Particularly, the quality of IoE services delivery must be maintained by analyzing contextual metrics of IoE such as people, data, process, and things. However, the challenges incorporate when the AI model conceives a lake of interpretation and intuition to the network service provider. Therefore, this paper provides an explainable artificial intelligence (XAI) framework for quality-aware IoE service delivery that enables both intelligence and interpretation. First, a problem of quality-aware IoE service delivery is formulated by taking into account network dynamics and contextual metrics of IoE, where the objective is to maximize the channel quality index (CQI) of each IoE service user. Second, a regression problem is devised to solve the formulated problem, where explainable coefficients of the contextual matrices are estimated by Shapley value interpretation. Third, the XAI-enabled quality-aware IoE service delivery algorithm is implemented by employing ensemble-based regression models for ensuring the interpretation of contextual relationships among the matrices to reconfigure network parameters. Finally, the experiment results show that the uplink improvement rate becomes $42.43\%$ and $16.32\%$ for the AdaBoost and Extra Trees, respectively, while the downlink improvement rate reaches up to $28.57\%$ and $14.29\%$. However, the AdaBoost-based approach cannot maintain the CQI of IoE service users. Therefore, the proposed Extra Trees-based regression model shows significant performance gain for mitigating the trade-off between accuracy and interpretability than other baselines.     

\end{abstract}

\section{INTRODUCTION}
In the era of technology transformation from fifth-generation (5G) to sixth-generation (6G) wireless networks, artificial intelligence (AI) \cite{IEEEhowto:Guo_XAI_1, IEEEhowto:Khan_6G_2, IEEEhowto:Hammadi_XAI_2, IEEEhowto:Munir_APNOMS_Int_Ser} becomes a key enabler to meet the requirements of the Internet of Everything (IoE) services \cite{IEEEhowto:Pant_IoE_3, IEEEhowto:Kang_IoE_1}. In case of the IoE, the quality of a service fulfillment not only depends on inter-connected physical objects (i.e., Internet of Things (IoT)), but it also relies on people, data, and process \cite{IEEEhowto:Pant_IoE_3, IEEEhowto:Kang_IoE_1}. Therefore, to enable high-quality \cite{IEEEhowto:3GPP_1} IoE services such as emergency and navigation \cite{IEEEhowto:Munir_PAN}, healthcare \cite{IEEEhowto:Hammadi_XAI_2}, energy \cite{IEEEhowto:Munir_APNOMS_Grid_Shepherd, IEEEhowto:Munir_TNSM_Meta}, and infotainment \cite{IEEEhowto:Khan_6G_2}, the utilization of contextual network and service metrics \cite{IEEEhowto:Data_set_5g} become essential. Thus, data-informed and AI-based techniques gain prominent attraction to ensure quality-aware IoE service delivery. However, interpretation and understanding by the service provider are important since intuitive reconfiguration is required to maintain a certain level of quality of IoE services. That quality can be evaluated by analyzing channel quality indicator (CQI) \cite{IEEEhowto:3GPP_1} from each IoE service user's feedback. Therefore, it is essential to develop such an AI framework that can analyze contextual metrics of each IoE service and interpret its findings to assure quality-aware IoE service delivery. 

In this work, we propose an explainable artificial intelligence (XAI)-enabled framework for quality-aware IoE service delivery, where coefficient for each contextual metric is estimated to understand the feature interpretation. In this framework, we have taken into account all four properties of each IoE service session such as people, data, process, and things (i.e., devices).   We consider user speed (i.e., people), download and upload processes, reference signal received power (RSRP), reference signal received quality (RSRQ), signal to interference and noise ratio (SINR) as contextual metrics of each IoE service session. To the best of our knowledge, this work provides one of the first XAI-enabled frameworks for IoE service delivery. We face several design challenges to developing the \emph{XAI-enabled framework} for quality-aware IoE service delivery:
\begin{itemize}
	\item First, how to differentiate among the contextual metrics with limited data of each IoE service, where the correlation of prominent features depends on permutation among them.
	
	\item Second, how to deal with user mobility (i.e., speed), in which both downlink and uplink rely excessively on the distance between the current position of each service user and next-generation NodeB (gNB).
		
	\item Third, how to ensure interpretation of each IoE service decision so that the service provider can proactively reconfigure in an autonomous manner with intuition for maintaining the quality of each IoE service user.
	
	\item  Finally, how to mitigate the trade-off between accuracy and interpretability when an AI algorithm is required for autonomous decision-making to capture network dynamics along with contextual metrics.
	
\end{itemize}

To address the aforementioned challenges, the main contribution of this work is summarized as follows: 
\begin{enumerate}
	\item First, we formulate a quality-aware IoE service delivery problem for a service provider by considering network dynamics and contextual metrics, where the objective is to maximize the channel quality index of each IoE service user. That can collectively maximize the quality of IoE service delivery performance.  
	
	\item Second, we propose an XAI-enabled framework for accomplishing quality-aware IoE service delivery to the users, where coefficients for the contribution of all contextual matrices are estimated by adopting well-known Shapley value \cite{IEEEhowto:Shapley_value_1} interpretation. As a result, the proposed framework is capable of deploying a variety of AI algorithms, as well as that can analyze and interpret contextual relationships among the matrices to reconfigure network parameters.
	
	\item Third, we develop an XAI-enabled quality-aware IoE service delivery algorithm for the proposed framework. Particularly, we solve the quality-aware IoE service delivery problem by implementing a Shapley-based regression model.
	
	\item Finally, we have performed rigorous experimental analysis by implementing ensemble-based \cite{IEEEhowto:Shapley_value_2, IEEEhowto:sklearn_ensemble} regression models such as Random Forest, Extra Trees, Gradient Boosting, AdaBoost, and Linear Regression are considered as XAI supported models, while Long short-term memory (LSTM), and deep neural network (DNN) are used as neural network models. We have found Extra Trees-based XAI can significantly better performance than others in terms of quality enhancement of IoE services.    
\end{enumerate}
We organize the rest of this paper as follows: Section \ref{system_model} presents the proposed system model and problem formulation of explainable artificial intelligence-enabled quality-aware IoE service delivery scheme. The proposed solution approach of the proposed XAI framework is discussed in Section \ref{sol_app}. Section \ref{Performance} demonstrates the performance analysis and key findings. Concluding remarks are given in Section \ref{Conclusion}. Additionally, Table \ref{sum_not} presents a summary of notations. 
\begin{table}[t!]
	\caption{Summary of Notations }
	\begin{center}
		\begin{tabular}{c|c}
			\hline
			\textbf{Description}&{\textbf{Notation}} \\
			\hline
			Channel Quality Indicator (CQI) &$\xi$ \\
			Downlink bit rate [Mbps] &$\Phi$ \\
			Internet of Everything (IoE) service user & $\mathcal{K} = \left\{{1,\dots,K}\right\}$\\
			Next generation NodeB (gNB)  & $\mathcal{B} = \left\{{1, \dots, B}\right\}$ \\
			Noise power [dBm] & $\lambda$\\
			Reference Signal Received Power (RSRP) [dBm] &$\alpha$ \\
			Reference Signal Received Quality (RSRQ) [dB] &$\beta$ \\
			RSSI [dBm] &$\mu$ \\
			Set of contextual matrices & $\mathcal{N}$\\
			Set of contextual coefficients & $\mathcal{C}$\\
			Signal to Interference \& Noise Ratio (SINR) [dB] &$\eta$ \\
			Speed [Km/h]&$\upsilon$ \\
			Uplink bit rate [Mbps] &$\Upsilon$ \\
			\hline
		\end{tabular}
		\label{sum_not}
	\end{center}
	\vspace{-4mm}
\end{table}

\section{System Model and Problem Formulation}
\label{system_model}
\begin{figure}[!t]
	\centerline{\includegraphics[width=\linewidth]{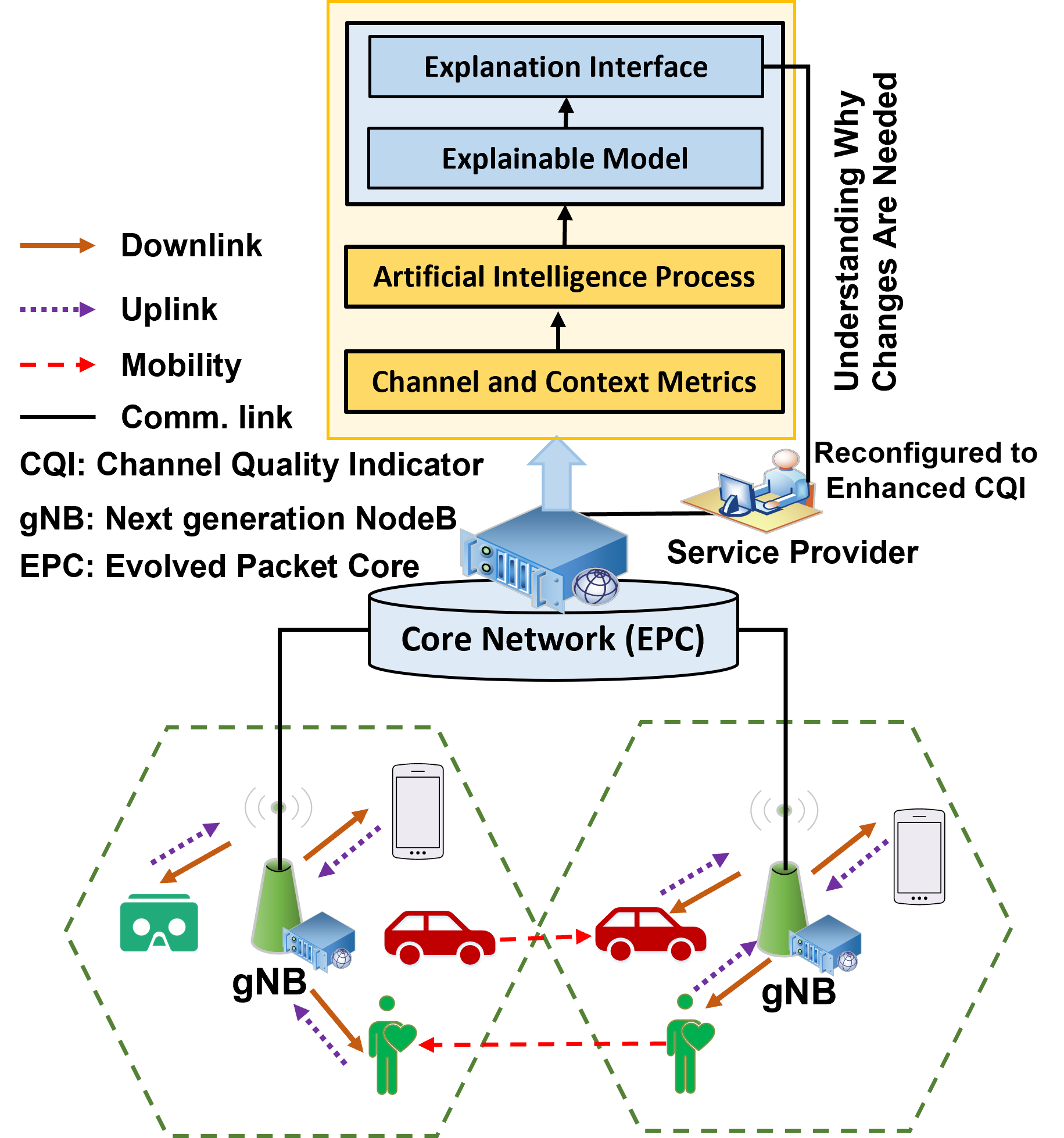}}
	\caption{A system model of explainable artificial intelligence-enabled quality-aware IoE service delivery scheme.}
	\label{System_model}
\end{figure}
Considering a wireless network (as shown in \ref{System_model}) that can support Internet of Everything (IoE) services for a certain area. Therefore, a set $\mathcal{B} = \left\{{1,2,\dots, B}\right\}$ of $B$ gNBs are physically deployed along with computational server. A set $\mathcal{K} = \left\{{1,2,\dots,K}\right\}$ of $K$ network users can take several network services \cite{IEEEhowto:Munir_IoT}, such as emergency, healthcare, navigation, and entertainment from the already subscribed service provider. Thus, various network entities are involved for ensuring quality-aware IoE service delivery to the users. In other words, these entities become IoE that includes people, data, processes, and things \cite{IEEEhowto:Kang_IoE_1, IEEEhowto:Pant_IoE_3}. In the considered system model, each gNB is connected with the core network, where the network service provider can monitor and reconfigure the network parameters for assuring a quality-aware service delivery to the users.

The quality of IoE service completely relies on the CQI $\xi_{k}$ feedback \cite{IEEEhowto:3GPP_1} by the service user (user equipment) $k \in \mathcal{K}$. While several contextual metrics, such as RSRP $\alpha_{k}$, RSRQ $\beta_{k}$, and SINR $\eta_{k}$ play key role for the IoE service quality. Additionally, the mobility (i.e., speed $\upsilon_{k}$) of user $k \in \mathcal{K}$ to/from the gNB $b \in \mathcal{B}$ becomes a crucial factor to ensure the quality IoE services. Thus, proper association $z_{k \rightarrow b} = 1$ of a service user $k \in \mathcal{K}$ with a gNB $b \in \mathcal{B}$ directly effects on data rate of both uplink $\Upsilon_{k}$ and downlink $\Phi_{k}$.     

\subsection{Quality-Aware IoE Service Delivery Model}
Let a service user $k \in \mathcal{K}$ receives RSSI $\mu_k$ form the nearby gNB $b \in \mathcal{B}$. In this system model, we consider $W = 20$ MHz wider range bandwidth based on orthogonal frequency-division multiplexing (OFDM), where each resource block (RB) contains $180$ KHz sub-carriers (i.e., $12$) and $100$ is the physical number of RBs at per channel bandwidth \cite{IEEEhowto:Data_set_5g, IEEEhowto:3GPP_1}. Therefore, reference signal received power (RSRP) of user $k \in \mathcal{K}$ can be estimated as follows \cite{IEEEhowto:3GPP_1}:
\begin{equation} \label{eq:RSRP}
	\begin{split}
		\alpha_{b \rightarrow k} = \mu_{b \rightarrow k} - 10 \log (12 \times J),
	\end{split}
\end{equation}
where $J$ denotes the number of physical resource block and $\mu_{b \rightarrow k}$ represents received RSSI by user $k \in \mathcal{K}$ from gNB $b \in \mathcal{B}$. Thus, we can capture reference signal received quality $\beta_{b \rightarrow k}$ of user $k$ for gNB $b$ as,
\begin{equation} \label{eq:RSRQ}
	\begin{split}
		\beta_{b \rightarrow k} = J \times \frac{\mu_{b \rightarrow k}}{\alpha_{b \rightarrow k}}.
	\end{split}
\end{equation}
For a noise power $\lambda_{k}$, we can estimate signal to interference plus noise ratio (SINR) of the user $k$ as follows \cite{IEEEhowto:Munir_TNSM_Meta, IEEEhowto:Munir_IoT}:
\begin{equation} \label{eq:SINR}
	\begin{split}
		\eta_{b \rightarrow k} = 10 \log \frac{\alpha_{b \rightarrow k}}{\lambda_{k} + \sum_{j \in \mathcal{B} \setminus b}I_{j}},
	\end{split}
\end{equation}
where $\alpha_{b \rightarrow k}$ represents RSRP at user device $k \in \mathcal{K}$. For a fixed bandwidth $W$, noise power $\lambda_{k}$ becomes $-174 + 10 \log (W)$ and $I_{j}$ denotes a transmission channel interference with other gNB in the considered system model. Thus, based on the current networks state, the user device $k \in \mathcal{K}$ sends channel quality indicator $\xi_{k \rightarrow b}$ to gNB $b \in \mathcal{B}$ as a feedback. We determine CQI as follows \cite{IEEEhowto:CQI_Li}:   
\begin{equation} \label{eq:CQI}
	\begin{split}
		\xi_{k \rightarrow b} =0.5223 \times \eta_{b \rightarrow k} + 4.6176,
	\end{split}
\end{equation}
where $\eta_{b \rightarrow k}$ is the SINR at user device $k \in \mathcal{K}$ for gNB $b \in \mathcal{B}$. 

\subsection{Problem Formulation of Quality-Aware IoE Service Delivery}
In this work, the objective is to maximize both the downlink and uplink data rate of the network for IoE service fulfillment. To do this, we need to improve the performance of each gNB $b \in \mathcal{B}$ so that it can accumulatively maximize the IoE service data rate. However, the challenges are to capture the dynamics of IoE by considering the contextual matrices of the network, such as RSRP, RSRQ, and SINR. Therefore, to dynamically adapt those contextual matrices, a data-informed scheme can be a suitable way. Thus, considering these contextual matrices, we can maximize the channel quality indicator of each IoE service user $k \in \mathcal{K}$, which can maximize the both uplink and downlink data rate of the network.
Considering a $\mathcal{N} = \left\{ \upsilon_{k}, \alpha_{k}, \beta_{k}, \eta_{k}, \mu_{k} \right\}$ set of contextual matrices for each user $k \in \mathcal{K}$, where $\upsilon_{k}$, $\alpha_{k}$, $\beta_{k}$, $\eta_{k}$, and $\mu_{k}$, represent speed, RSRP, RSRQ, SINR, and RSSI, respectively. Therefore, using \eqref{eq:SINR} and \eqref{eq:CQI}, the quality-aware IoE service delivery model can represent as follows:
\begin{equation} \label{eq:Obj}
	\begin{split}
		\Lambda (\mathcal{N})  = 0.5223 \times \big( 10 \log \frac{\alpha_{b \rightarrow k}}{\lambda_{k} + \sum_{j \in \mathcal{B} \setminus b}I_{j}} \big) + 4.6176.
	\end{split}
\end{equation}
In \eqref{eq:Obj}, the quality of each IoE service completely relies on correlation among $\upsilon_{k}$, $\alpha_{k}$, $\beta_{k}$, $\eta_{k}$, and $\mu_{k}$. Thus, we consider $C$ is the coalition among them (i.e., features) as well as  downlink $\Phi_{k}$, and uplink $\Upsilon_{k}$ data rate \cite{IEEEhowto:Shapley_value_1, IEEEhowto:Shapley_value_2}. Theretofore, we need to adjust $\mathcal{C}$ during optimization to maximize the CQI of each user $k \in \mathcal{K}$, where $\left\{ \upsilon_{k}, \alpha_{k}, \beta_{k}, \eta_{k}, \mu_{k}, \xi_{k}, \Phi_{k}, \Upsilon_{k} \right\} \in \mathcal{C} $. In which, CQI $\xi_{k}$ by an user $k \in \mathcal{K}$ depends on the achieved bit rate of both downlink $\Phi_{k}$, and uplink $\Upsilon_{k}$. As a result, we can rewrite the set of contextual matrices for each user $k \in \mathcal{K}$ as, $\mathcal{N} = \left\{ \upsilon_{k}, \alpha_{k}, \beta_{k}, \eta_{k}, \mu_{k}, \xi_{k}, \Phi_{k}, \Upsilon_{k} \right\}$. Then, we can formulate the quality-aware IoE service delivery problem as follows:
\begin{subequations}\label{Opt_1_1}
	\begin{align}
		\underset{\boldsymbol{z}, \boldsymbol{\Phi}, \boldsymbol{\Upsilon}}\max \sum_{b \in \mathcal{B}} 
		&\;  \sum_{k \in \mathcal{K}} \mathcal{C} \Lambda (\mathcal{N})  \tag{\ref{Opt_1_1}}, \\
		\text{s.t.} \quad & \label{Opt_1_1:const1} z_{k \rightarrow b} \alpha_{b \rightarrow k} \ge \omega,  \alpha_{b \rightarrow k} \in  \mathcal{N},\\
		&\label{Opt_1_1:const2} z_{k \rightarrow b} \beta_{b \rightarrow k} \ge \zeta, \beta_{b \rightarrow k} \in  \mathcal{N},\\
		&\label{Opt_1_1:const3}  z_{k \rightarrow b}\upsilon_{k} \times \Delta t_{k \rightarrow b}  \times 1000 \ge  h^{\textrm{max}},\\
		&\label{Opt_1_1:const4} \mathcal{C} \subseteq  \mathcal{N}, \mathcal{C} \in 2^N \rightarrow \mathbb{R}, \\
		& \label{Opt_1_1:const5} z_{k \rightarrow b} \in \left\lbrace0,1 \right\rbrace,  \forall k \in \mathcal{K}.
	\end{align}
\end{subequations}
In quality-aware IoE service delivery problem \eqref{Opt_1_1}, $\forall z_{k \rightarrow b} \in \boldsymbol{z}$, $\forall \Phi_{b \rightarrow k} \in \boldsymbol{\Phi}$, and  $\forall \Upsilon_{k \rightarrow b} \in \boldsymbol{\Upsilon}$ are the decision variables. $z_{k \rightarrow b} \in \boldsymbol{z}$ represents an association variable of user $k \in \mathcal{K}$ to gNB $b \in \mathcal{B}$, $\Phi_{b \rightarrow k} \in \boldsymbol{\Phi}$ denotes downlink data rate of user $k \in \mathcal{K}$ from gNB $b \in \mathcal{B}$, and $\Upsilon_{k \rightarrow b} \in \boldsymbol{\Upsilon}$ is the uplink data rate of user $k \in \mathcal{K}$ to gNB $b \in \mathcal{B}$. 
Constraint \eqref{Opt_1_1:const1} ensures reference signal received power must be satisfied a minimum level of RSRP $\omega$. Similarly, constraint  \eqref{Opt_1_1:const2} assures a certain level $\zeta$ of reference signal received quality for  $k \in \mathcal{K}$ by gNB $b \in \mathcal{B}$. Mobility (i.e., speed) of each IoE service user $k \in \mathcal{K}$ is taken into account in constraint \eqref{Opt_1_1:const3} of the formulated problem \eqref{Opt_1_1}. Where $\Delta t_{k \rightarrow b}$ denotes the changes of time towards gNB $b \in \mathcal{B}$. A correlation among the contextual matrices (i.e., features) are established in constraint \eqref{Opt_1_1:const4}, where $C$ represent the coalition among contextual features of IoE service delivery matrices. Finally, constraint \ref{Opt_1_1:const5} ensures each IoE service user $k \in \mathcal{K}$ with in the signal range of gNB $b \in \mathcal{B}$.  Decisions of the formulated problem \eqref{Opt_1_1} leads to $\mathcal{O}(2^{|\mathcal{B}|\times|\mathcal{K}| \times |\mathcal{N}| })$, where $|\mathcal{B}|$, $|\mathcal{K}|$, and $|\mathcal{N}|$ are the number of gNB, user, and features, respectively. We can obtain a global optimal solution when complexity grows exponentially \cite{IEEEhowto:Munir_PAN, IEEEhowto:Munir_IoT}. Thus, hard to solve the formulated problem \eqref{Opt_1_1} into polynomial time. Therefore, we design an approximate solution by modeling a multi-variant regression \cite{IEEEhowto:Shapley_value_2} problem based on explainable artificial intelligence (XAI) model. The contextual features are taken into account by forming a coalitional game among contextual matrices \cite{IEEEhowto:Shapley_value_1, IEEEhowto:Shapley_value_2}. So that the service provider can get the answer of the question, \emph{why changes are required to enhance CQI of each IoE user $k \in \mathcal{K}$?}. A detailed solution design is presented in the following section.

\section{XAI-enabled IoE Service Delivery Framework}
\label{sol_app}
Recall the set $\mathcal{N} = \left\{ \upsilon_{k}, \alpha_{k}, \beta_{k}, \eta_{k}, \mu_{k}, \xi_{k}, \Phi_{k}, \Upsilon_{k} \right\}$ of $N$ contextual matrices, where $\mathcal{N}$ consist of $|\mathcal{N}|$ players and indexed by $i$. Thus, a characteristic function $\varphi$ can map all subset of $\mathcal{N}$ contextual features to $\varphi \colon 2^{|\mathcal{N}|} \rightarrow \mathbb{R}$. If feature $i \in \mathcal{N}$ form a coalition $\mathcal{C}$ with other players $\mathcal{C} \subseteq \mathcal{N} \setminus \left\{ i \right\}$ then for a coalition game $(\varphi, \mathcal{N})$ the Shapley value can be calculated as follows \cite{IEEEhowto:Shapley_value_1, IEEEhowto:Shapley_value_2}: 
\begin{equation} \label{eq:shap_val}
	\begin{split}
		\Psi_i (\varphi) = \sum_{\mathcal{C} \subseteq \mathcal{N} \setminus \left\{ i \right\}} \frac{|\mathcal{C}|! (|\mathcal{N}|- |\mathcal{C}|-1)!}{|\mathcal{N}|!} (\varphi (\mathcal{C} \cup \left\{i\right\}) - \varphi(\mathcal{C})),
	\end{split}
\end{equation}
where $|\mathcal{N}|$ represents number of contextual matrices (i.e., players) and for player $i$, $(\varphi (\mathcal{C} \cup \left\{i\right\}) - \varphi(\mathcal{C}))$ denotes contribution of a fair compensation. Additionally, $\varphi(\mathcal{C})$ is the worth of coalition $\mathcal{C}$. Therefore, intuitively we can rewrite \eqref{eq:shap_val} as follows \cite{IEEEhowto:Shapley_value_1, IEEEhowto:Shapley_value_2}:
\begin{equation} \label{eq:shap_interpretation}
	\begin{split}
		\Psi_i (\varphi) = \frac{1}{|\mathcal{N}|}\sum_{\mathcal{C} \subseteq  \mathcal{N} \setminus \left\{ i \right\}} \frac{marginal \; contribution \; of \; i}{number \; of \; coalitions \; \mathcal{C} \subseteq \mathcal{N} \setminus \left\{ i \right\} } \\= \frac{1}{|\mathcal{N}|}\sum_{\mathcal{C} \subseteq  \mathcal{N} \setminus \left\{ i \right\}} \left( \begin{array}{c} |\mathcal{N}|-1 \\|\mathcal{C}| \end{array} \right)^{-1}(\varphi (\mathcal{C} \cup \left\{i\right\}) - \varphi(\mathcal{C})).
	\end{split}
\end{equation}
Let consider coefficients for contribution of all contextual matrices are represented a set of $\mathcal{X} \coloneqq \forall \boldsymbol{x}$ such that $\boldsymbol{x} = \varphi (\left\{ i, \dots, N \right\})$. Therefore, the model coefficients are calculated as follows:   
\begin{equation} \label{eq:shap_val_Dis}
	\begin{split}
		\boldsymbol{x} = \Psi_{i} (\varphi), \dots, \Psi_{N} (\varphi), \forall i \in \mathcal{N}, \forall \boldsymbol{x} \in \mathcal{X}.
	\end{split}
\end{equation}

In this solution, our aim is to design a XAI-based multi-variant regression model to solve the quality-aware IoE service delivery problem \eqref{Opt_1_1}. Particularly, we consider Shapley value coefficients to interpret contextual relationship among the matrices so that IoE service provider can reconfigured network parameters for meeting the service quality (i.e., CQI) of all users $\forall k \in \mathcal{K}$. Thus, coefficients of contextual features $(\Psi_{i} (\varphi), \dots, \Psi_{N}(\varphi))$ is estimated as follows \cite{IEEEhowto:Shapley_value_2}:
\begin{equation} \label{eq:Regression}
	\begin{split}
		\boldsymbol{Z}_k =  \epsilon + \Psi_{i} (\varphi)i_{k} + \dots + \Psi_{N} (\varphi)N_{k}, \forall i \in \mathcal{N},
	\end{split}
\end{equation}
where $i_k, \dots N_k$ represent the contextual features of IoE service for user $k \in \mathcal{K}$. The training loss function of the quality-aware IoE service delivery model is as follows:   
\begin{equation} \label{eq:Regression_Cost}
	\begin{split}
		\mathbb{E}(\epsilon, \mathcal{N}) = 	\underset{\boldsymbol{z}, \boldsymbol{\Phi}, \boldsymbol{\Upsilon}}\min \frac{1}{2|\mathcal{K}|} \sum_{k=1}^{|\mathcal{K}|} (\Lambda (\mathcal{N}) - \boldsymbol{Z}_k), \forall i \in \mathcal{N},
	\end{split}
\end{equation}
where $\epsilon$ denotes the intercept, $\mathcal{N}$ represents contextual input (i.e., $\left\{i, \dots, N \right\}$), while $\boldsymbol{z}$, $\boldsymbol{\Phi}$, and $\boldsymbol{\Upsilon}$ are the decision variables. In \eqref{eq:Regression_Cost}, the objective is to minimize the loss while considering the Shapley value coefficients for contextual interpretation. We present the overall algorithmic procedure in Algorithm \ref{alg:XAI_algo}.   

The service provider is responsible for deploying the quality-aware IoE service delivery Algorithm \ref{alg:XAI_algo} at evolved packet core (EPC) (as seen in Figure \ref{System_model}). The input set $\mathcal{N}, \forall k \in \mathcal{K}$ of Algorithm \ref{alg:XAI_algo} consists of user speed $\upsilon_{k}$, RSRP $\alpha_{k}$, RSRQ $\beta_{k}$, SINR $\eta_{k}$, RSSI $\mu_{k}$,  CQI $\xi_{k}$, downlink bit rate $\Phi_{k}$, and uplink bit rate $\Upsilon_{k}$ form the historical data. While the output of the Algorithm \ref{alg:XAI_algo} includes association  $\forall z_{k \rightarrow b} \in \boldsymbol{z}$ for all user $\forall k \in \mathcal{K}$ to one gNB $b \in \mathcal{B}$, downlink data rate $\forall \Phi_{b \rightarrow k} \in \boldsymbol{\Phi}$, and uplink data rate $\forall \Upsilon_{k \rightarrow b} \in \boldsymbol{\Upsilon}$. Additionally, contextual coefficients $\boldsymbol{x} \eqqcolon (\Psi_{i} (\varphi), \dots, \Psi_{N}(\varphi)), \forall \boldsymbol{x} \in \mathcal{X}$ are also estimating in Algorithm \ref{alg:XAI_algo} to reconfigure network parameters $\mathcal{N}$ for ensuring quality of IoE service delivery. In Algorithm \ref{alg:XAI_algo}, lines from $3$ to $7$ calculate RSRP, RSRQ, SINR, and CQI for IoE user $k \in \mathcal{K}$, where based on received RSRP from gNB $b \in \mathcal{B}$, user $k \in \mathcal{K}$ is associated with gNB $b \in \mathcal{B}$ in line $4$. Lines from $8$ to $13$ are responsible for finding subsets of all contextual features (in Algorithm \ref{alg:XAI_algo}). Particularly, a coalition is form to find the contextual coefficient of each feature in line $10$ of Algorithm \ref{alg:XAI_algo}. In line $15$, the coefficients of contextual features $(\Psi_{i} (\varphi), \dots, \Psi_{N}(\varphi))$ are estimated based on a multi-variant regression model. Finally, explainable AI-based model for quality-aware IoE service delivery is trained in line $18$ of Algorithm \ref{alg:XAI_algo}. The complexity of Algorithm \ref{alg:XAI_algo} leads to $\mathcal{O}(|\mathcal{B}| \times|\mathcal{K}| \times |\mathcal{N}| + 2^{|\mathcal{N}|})$. Detailed performance evaluation and discussion of the proposed XAI-based quality-aware IoE service delivery of Algorithm \ref{alg:XAI_algo} are given in the following section.  
\begin{algorithm}[!t]
	\caption{XAI-enabled Quality-Aware IoE Service Delivery}
	\label{alg:XAI_algo}
	\begin{algorithmic}[1]
		\renewcommand{\algorithmicrequire}{\textbf{Input:}}
		\renewcommand{\algorithmicensure}{\textbf{Output:}}
		\REQUIRE $\mathcal{N} = \left\{ \upsilon_{k}, \alpha_{k}, \beta_{k}, \eta_{k}, \mu_{k}, \xi_{k}, \Phi_{k}, \Upsilon_{k} \right\}$, $\forall k \in \mathcal{K}$
		\ENSURE  $\boldsymbol{z}$, $\boldsymbol{\Phi}$, $\boldsymbol{\Upsilon}$, $\mathcal{X}$  
		\\ \textbf{Initialization}: $J$, $\mathcal{X}$, $\mathcal{C}$, $\epsilon$
		\WHILE{$\forall b \in \mathcal{B}$} 
		\FOR {$\forall k \in \mathcal{K}$ }
		\STATE \textbf{Calculate current RSRP:} $\alpha_{b \rightarrow k}$ using \eqref{eq:RSRP}
		\STATE \textbf{Assign:} $z_{k \rightarrow b} = 1$
		\STATE \textbf{Calculate current RSRQ:} $\beta_{b \rightarrow k}$, using \eqref{eq:RSRQ}.
		\STATE \textbf{Estimate SINR:} $\eta_{b \rightarrow k}$ using \eqref{eq:SINR}
		\STATE \textbf{Get current CQI:} $\xi_{k \rightarrow b}$ using \eqref{eq:CQI}
		\FOR {$\forall i \in \mathcal{N}$ }
		\IF {$\mathcal{C} \subseteq \mathcal{N} \setminus \left\{ i \right\}$}
		\STATE\textbf{Calculate:} $\Psi_i (\varphi)=$\\ $\frac{1}{|\mathcal{N}|}\sum_{\mathcal{C} \subseteq \mathcal{N} \setminus  \left\{ i \right\}} \left( \begin{array}{c} |\mathcal{N}|-1 \\|\mathcal{C}| \end{array} \right)^{-1}(\varphi (\mathcal{C} \cup \left\{i\right\}) - \varphi(\mathcal{C}))$ using \eqref{eq:shap_interpretation}
		\STATE\textbf{Append:} $\Psi_i (\varphi) \in \boldsymbol{x}$  
		\ENDIF
		\ENDFOR
		\STATE\textbf{Append:} $\mathcal{X} \coloneqq \forall \boldsymbol{x}$ using \eqref{eq:shap_val_Dis} \STATE\textbf{Estimate:} $(\Psi_{i} (\varphi), \dots, \Psi_{N}(\varphi))$ using  \eqref{eq:Regression}
		\ENDFOR
		\IF {$z_{k \rightarrow b}$ == 1}
		\STATE\textbf{Evaluate:} $	\underset{ \boldsymbol{z}, \boldsymbol{\Phi}, \boldsymbol{\Upsilon} }\min \frac{1}{2|\mathcal{K}|} \sum_{k=1}^{|\mathcal{K}|} (\Lambda (\mathcal{N}) - \boldsymbol{Z}_k)$ using \eqref{eq:Regression_Cost}
		\ENDIF
		\ENDWHILE
		\STATE\textbf{Calculate:} $\forall \Phi_{b \rightarrow k} \in \boldsymbol{\Phi}$, and  $\forall \Upsilon_{k \rightarrow b} \in \boldsymbol{\Upsilon}$
		\STATE\textbf{Configure:} $\mathcal{X}$ to gNB $\boldsymbol{z}$, $\forall z_{k \rightarrow b} \in \boldsymbol{z}$
		\RETURN  $\boldsymbol{z}$, $\boldsymbol{\Phi}$, $\boldsymbol{\Upsilon}$ 
	\end{algorithmic} 
\end{algorithm}

\section{Performance Evaluation and Discussion}
\label{Performance}
\begin{table}[t!]
	\caption{Summary of Experiment Setup }
	\begin{center}
		\begin{tabular}{c|c}
			\hline
			\textbf{Simulation Parameters}&{\textbf{Values}} \\
			\hline
			No. of gNB $|\mathcal{B}|$ & $8$ \cite{IEEEhowto:Data_set_5g}\\ 
			No. of user sessions $|\mathcal{K}|$ [train, test] & $[1544, 662]$ \cite{IEEEhowto:Data_set_5g}\\
			Max. speed & $88$ [Km/h] \cite{IEEEhowto:Data_set_5g}\\
			Max. downlink data rate & $170.06$ [Mbps] \cite{IEEEhowto:Data_set_5g}\\
			Max. uplink data rate & $0.825$ [Mbps] \cite{IEEEhowto:Data_set_5g}\\
			Download file size& $>200$ MB \cite{IEEEhowto:Data_set_5g} \\
			Bandwidth [wider range] & $20$ MHz \cite{IEEEhowto:3GPP_1}\\
			No. of sub-carriers  & $12$ \cite{IEEEhowto:3GPP_1}\\
			Each resource block carrier frequency  & $180$ KHz \cite{IEEEhowto:3GPP_1}\\
			Physical no. of RBs at per channel BW  & $100$ \cite{IEEEhowto:3GPP_1}\\
			No. of estimator [AdaBoost \& Extra Trees]  & $100$ \\
			No. of training epochs [LSTM \& DNN]  & $500$ \\
			Batch size [LSTM \& DNN]  & $72$ \\
			No. of LSTM units  & $100$ \\
			No. of dense layers [DNN]  & $100$ \\
			\hline	
		\end{tabular}
		\label{tab2_sim_param}
	\end{center}
	\vspace{-4mm}
\end{table}
\begin{figure}[!t]
\centerline{\includegraphics[scale=0.55]{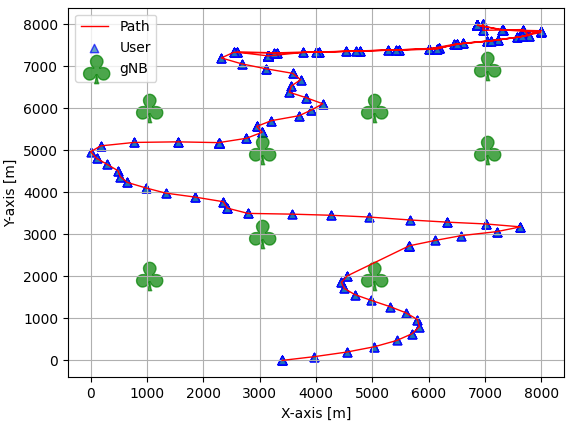}}
\caption{Considered topology for evaluating the proposed XAI-enabled IoE service delivery framework based on dataset (B$\_2020.02.13\_13.03.24$) \cite{IEEEhowto:Data_set_5g}.}
\label{topology}
\end{figure}
We have evaluated the effectiveness of the proposed explainable artificial intelligence framework using a state-of-the-art $5$G dataset (B$\_2020.02.13\_13.03.24$) \cite{IEEEhowto:Data_set_5g}. We have shown the considered topology in Figure \ref{topology} and the important parameters are in Table \ref{tab2_sim_param}. We consider a desktop computer with a Core i$9$ processor ($2.8$ GHz) along with $64$ GB of random access memory as an EPC core computational server to execute the implemented Algorithm \ref{alg:XAI_algo}. The implementation and scientific evaluation has been done on top of the Python framework \cite{IEEEhowto:sklearn_ensemble, IEEEhowto:SHAP}. We have implemented numerous ensemble-based regression schemes \cite{IEEEhowto:sklearn_ensemble} such as Random Forest, Extra Trees, Gradient Boosting, AdaBoost, and Linear Regression as XAI-supported models. Further, we consider LSTM and DNN-based regression algorithms like neural networks model for a fair comparison in terms of interoperability of AI models. We consider the ground truth of the dataset (B$\_2020.02.13\_13.03.24$) \cite{IEEEhowto:Data_set_5g} as the theoretical baseline. 
\begin{figure}[!t]
	\centerline{\includegraphics[width=\linewidth]{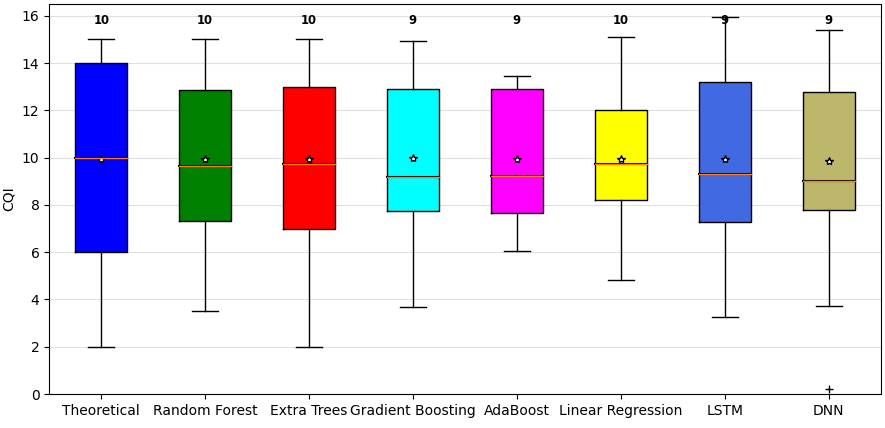}}
	\caption{A comparison of quality measurement during IoE service delivery.}
	\label{CQI_Box_round}
	\vspace{-6mm}
\end{figure}

\begin{figure}[!t]
	\centerline{\includegraphics[width=\linewidth]{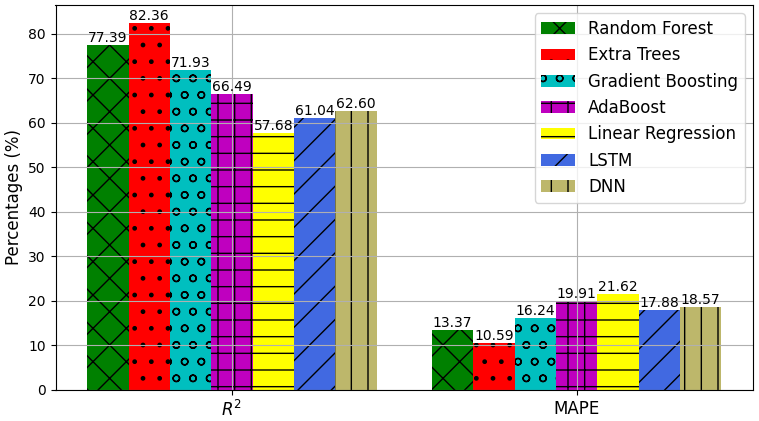}}
	\caption{R-squared ($R^2$) score and mean absolute percentage error (MAPE) analysis based on numerous regression models of the proposed quality-aware IoE service delivery.}
	\label{Score}
\end{figure}

\begin{figure}[!t]
	\centerline{\includegraphics[width=\linewidth]{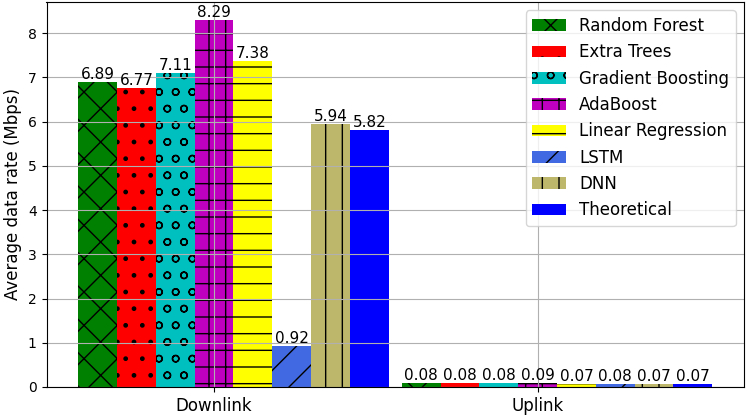}}
	\caption{A comparison of the achieved downlink and uplink data rate for the quality-aware IoE service delivery based on numerous regression models during execution (i.e., testing).}
	\label{data_rate}
	\vspace{-6mm}
\end{figure}

A comparison of quality measurement (i.e., CQI) for IoE service delivery (i.e., $662$ sessions) is illustrated in Figure \ref{CQI_Box_round}. Due to the interpretability of XAI-based models (i.e., Random Forest, Extra Trees, Linear Regression) can achieve an average CQI score of $10$ as the same as theoretical while neural networks-based LSTM and DNN can get $9$ in an average with a lower range percentile. Because the proposed XAI framework can control Shapley value-based prominent features (coefficients) during model training based on contextual metrics of IoE service. Similarly, higher R-squared ($R^2$) scores and lower mean absolute percentage error (MAPE) of XAI-based models (in Figure \ref{Score}) demonstrate the effectiveness of the Shapley coefficient \cite{IEEEhowto:Shapley_value_2} for AI model training. 

Further, a comparison of the achieved downlink and uplink data rate for the quality-aware IoE service delivery based on numerous regression models is presented in Figure \ref{data_rate}. We consider the relative improvement rate as a comparison metric based on the reference value (i.e., $improvement \;(\%) = ((actual \; improvement / reference \; value) \times 100) $). On the one hand, in Figure \ref{data_rate}, the uplink improvement rate of the AdaBoost ($0.09$ Mbps) and the Extra Trees ($0.08$ Mbps) based on theoretical ($0.07$ Mbps) are $28.57\%$ and $14.29\%$, respectively. On the other hand, the downlink gain rate of the AdaBoost ($8.29$ Mbps) and the Extra Trees ($6.77$ Mbps) become $42.43\%$ and $16.32\%$, respectively, as compared to theoretical ($5.82$ Mbps) measure. Although, the AdaBoost performs better with respect to the improvement of uplink and downlink data rate (in Figure \ref{data_rate}); however, the average CQI (in Figure \ref{CQI_Box_round}) and R-squired score (in Figure \ref{Score}) are $11.11\%$ and $19.87\%$ lower than that the Extra Trees-based XAI model. Therefore, by considering the trade-offs among the CQI, R-squired score, and both data rates (i.e., uplink and downlink), we proposed an Extra Trees-based XAI scheme as a solution of quality-aware IoE service delivery for the next-generation wireless networks.
Note that, the LSTM cannot predict properly the downlink data rate for IoE services as seen in Figure \ref{data_rate}. Thus, the LSTM is unable to discretize the significant feature during testing although $43,701$ parameters are used during the training. This is because of sensitivity to its random weight initialization, exploding, and vanishing gradient during training since a huge amount of variation among the downlink data rate of each IoE service request in the dataset (B$\_2020.02.13\_13.03.24$) \cite{IEEEhowto:Data_set_5g}. Therefore, training with high variation and a small amount of data LSTM cannot predict properly the downlink data rate of IoE services.
Other models can perform better in terms of achieved downlink and uplink data rate of IoE services.  
\begin{figure}[!t]
	\centerline{\includegraphics[width=\linewidth]{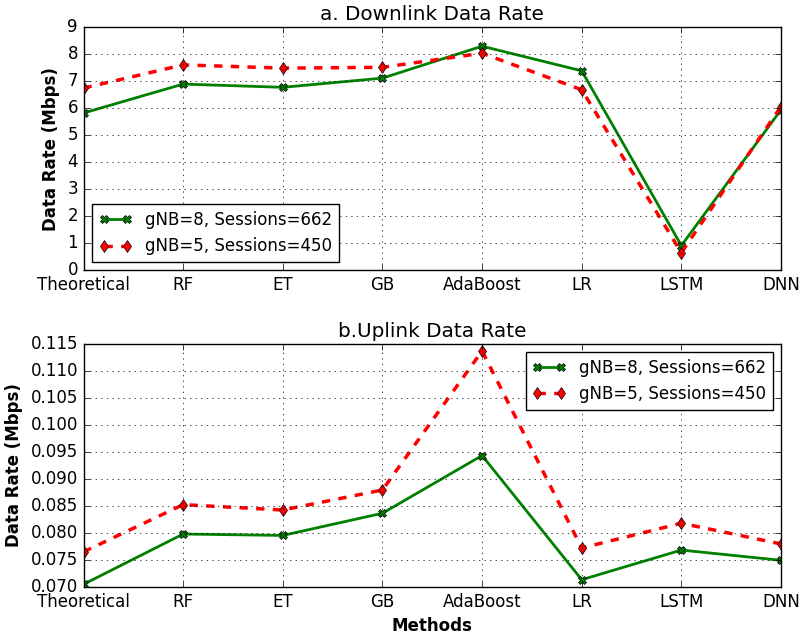}}
	\caption{The trend analysis of the proposed XAI framework when the parameters such as the number and placement of gNBs, and the number of user sessions are changed.}
	\label{Down_Up_Sessions_662_450}
\end{figure}

In Figure \ref{Down_Up_Sessions_662_450}, we have analyzed the trend of the experiment results (i.e., downlink and uplink data rate) of the proposed XAI framework when the parameters, such as the number and placement of gNBs, and the number of user sessions are changed. In particular, we have analyzed by comparing $8$ gNBs and $662$ sessions along with $5$ gNBs and $450$ user sessions in Figure \ref{Down_Up_Sessions_662_450}. This analogy ensures the effectiveness of the proposed XAI framework by maintaining the same trend of outcomes even if the network topology and environment have changed.

\begin{figure}[!t]
	\centerline{\includegraphics[width=\linewidth]{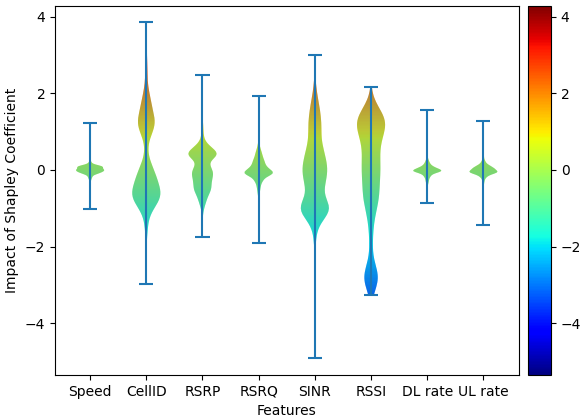}}
	\caption{Interpretation of the impact of Shapley value coefficient among the contextual features during AI model training and execution.}
	\label{Shap_coefficient}
\end{figure}
 
\begin{figure}[!t]
		\centerline{\includegraphics[width=8.3cm]{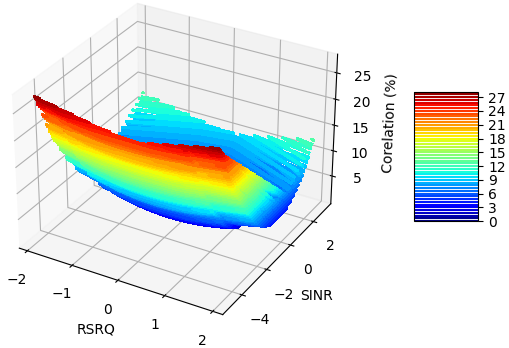}}
	\caption{Interpretation of the correlation between SINR and RPRQ based on contextual coefficient among them.}
	\label{SNR_RPRQ}
	\vspace{-6mm}
\end{figure}

Figure \ref{Shap_coefficient} illustrates the interpretation of Shapley value coefficient impact among the contextual features during AI model training and execution for the IoE service delivery. Figure \ref{Shap_coefficient} demonstrates that the proposed XAI framework has found RSSI, SINR, RSRP, and Cell ID (i.e., location of gNB) have more effect than that the other contextual features for enhancing CQI during the IoE service delivery to the users. Therefore, this analogy is quite practical and intuitive for the service provider to reconfigure network parameters for maintaining a certain level of quality of IoE service delivery. Note that, by characterizing the Shapley value coefficient (in \eqref{eq:shap_interpretation}) of the contextual features, the XAI-supported loss function \eqref{eq:Regression_Cost} minimizes its loss during the learning process while it fulfills the goal of \eqref{Opt_1_1} by maximizing the CQI of \eqref{eq:CQI}. Finally, Figure \ref{SNR_RPRQ} interprets an explanation and correlation between two major network metrics (i.e., RSRQ and SINR) that have a positive effect to enhance the CQI of IoE service users. 
In particular, we have found that a $27\%$ of correlation between SINR and RSRQ, where RSRQ depends on RSSI and RSRP (in \eqref{eq:RSRQ}) and SINR strongly relies on RSRP \eqref{eq:SINR}. Therefore, theoretically and intuitively, the proposed XAI framework can perfectly interpret the root cause of the AI decisions for quality-aware IoE service delivery. Additionally, the proposed XAI framework can flexibly reconfigure the network service parameters based on the contextual features of service requirements, for instance, it can prioritize the emergency IoE services over others by analyzing the contextual coefficient.

\section{Conclusion}
\label{Conclusion}
In this work, we are enabling a quality-aware IoE service delivery mechanism by proposing a new explainable artificial intelligence framework for EPC core. As a result, the IoE service provide can analyze and interpret contextual relationships among the features to reconfigure network parameters for maintaining a certain level of CQI of the users. In particular, this work is introducing an XAI framework that can flexibly incorporate several AI models based on service providers' requirements for autonomous control and interpretation of contextual metrics based on Shapley coefficients. The experimental results show that the proposed Extra Trees-based XAI regression model can enhance a significant amount of downlink $16.32\%$, and uplink $14.29\%$ data rate in terms of IoE service delivery than that the baseline while maintain the CQI. In the future, we will incorporate distributed scheme for analyzing contextual coefficients among the feature multiple service providers.

\vspace{12pt}

\end{document}